\documentclass{article}

\usepackage[final,nonatbib]{neurips_2019_ml4ad}

\usepackage[utf8]{inputenc} % allow utf-8 input
\usepackage[T1]{fontenc}    % use 8-bit T1 fonts
\usepackage{hyperref}       % hyperlinks
\usepackage{url}            % simple URL typesetting
\usepackage{booktabs}       % professional-quality tables
\usepackage{amsfonts}       % blackboard math symbols
\usepackage{nicefrac}       % compact symbols for 1/2, etc.
\usepackage{microtype}      % microtypography

\usepackage{xcolor}
\usepackage{subcaption}
\usepackage{graphicx}
\usepackage{multirow}
\usepackage{colortbl}
\usepackage{hhline}
\usepackage{enumitem}
\usepackage{verbatim}
\usepackage{float}

\title{RST-MODNet: Real-time Spatio-temporal Moving Object Detection for Autonomous Driving}

\author{%
  Mohamed Ramzy$^1$, Hazem Rashed$^2$, Ahmad El Sallab$^2$ and Senthil Yogamani$^3$ \\
  $^1$Cairo University $^2$Valeo R\&D, Egypt $^3$Valeo Vision Systems, Ireland \\
  \texttt{mohamed.ibrahim98@eng-st.cu.edu.eg} \\ \texttt{\{hazem.rashed,ahmad.el-sallab,senthil.yogamani\}@valeo.com} 
}

\begin{document}
\maketitle
\begin{abstract} \label{sec:abstract}
     Moving Object Detection (MOD) is a critical task for autonomous vehicles as moving objects represent higher collision risk than static ones.  The trajectory of the ego-vehicle is planned based on the future states of detected moving objects. It is quite challenging as the ego-motion has to be modelled and compensated to be able to understand the motion of the surrounding objects. 
     In this work, we propose a real-time end-to-end CNN architecture for MOD utilizing spatio-temporal context to improve robustness. We construct a novel time-aware architecture exploiting temporal motion information embedded within sequential images in addition to explicit motion maps using optical flow images.
     We demonstrate the impact of our algorithm on KITTI dataset where we obtain an improvement of 8\% relative to the baselines. We compare our algorithm with state-of-the-art methods and achieve competitive results on KITTI-Motion dataset in terms of accuracy at three times better run-time. The proposed algorithm runs at 23 fps on a standard desktop GPU targeting deployment on embedded platforms.
\end{abstract}
\section{Introduction} \label{sec:introduction}

\begin{figure}
  \centering
%   \fbox{\rule[-.5cm]{0cm}{4cm} \rule[-.5cm]{4cm}{0cm}}
\includegraphics[width=\textwidth]{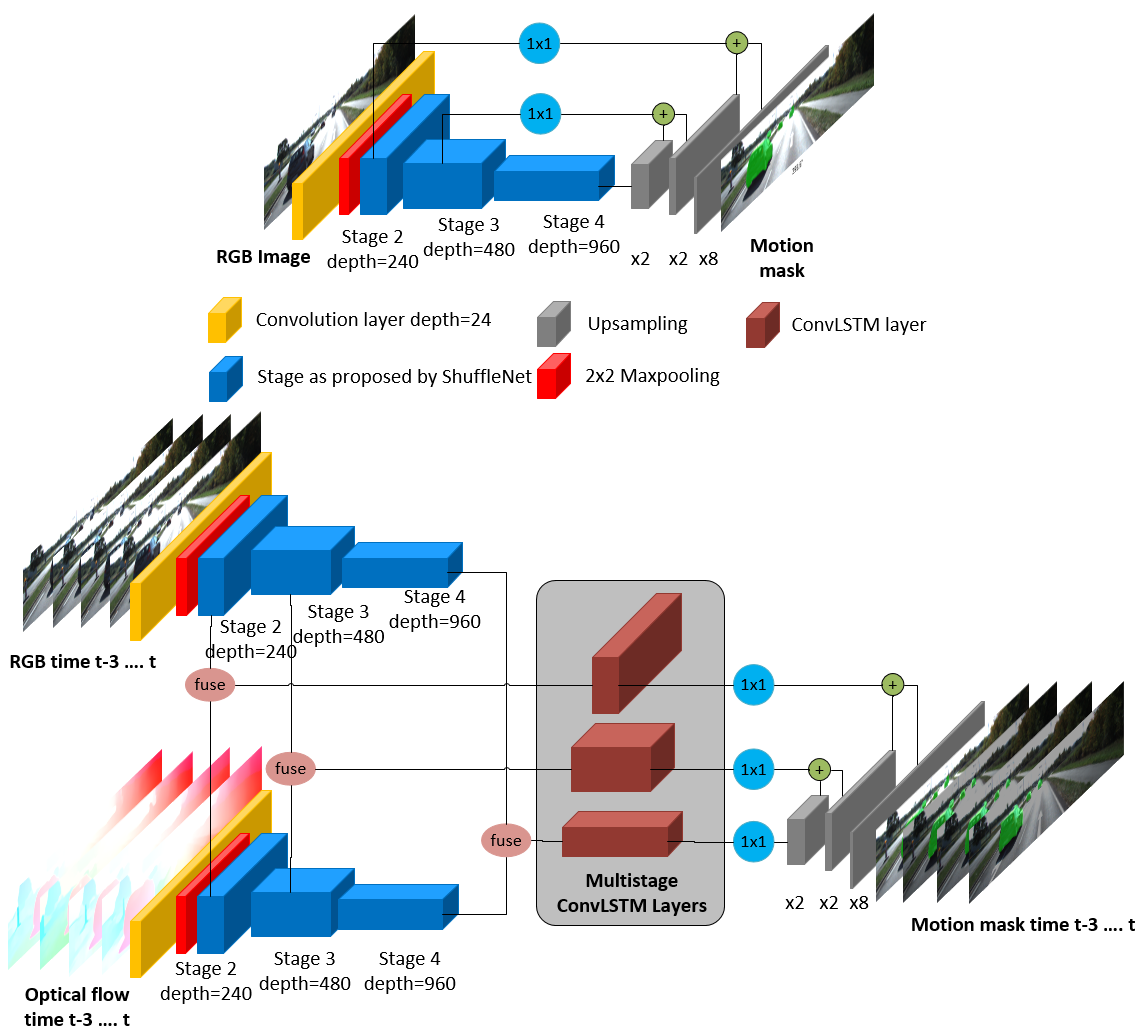}
%     \vspace{-1cm}
  \caption{\textbf{Top:} Our baseline architecture based on ShuffleNet \cite{zhang2018shufflenet}. \textbf{Bottom: } Our proposed architecture after optical flow and time augmentation.}
  \label{fig:arch}
\end{figure}

% \vspace{-1cm}

The Autonomous Driving (AD) scenes are highly dynamic as they contain multiple object classes that move at different speeds in diverse directions \cite{horgan2015vision, heimberger2017computer}. It is critical to understand the motion model of each of the surrounding elements for the purpose of planning the ego-trajectories considering the future positions and velocities of these objects. Typically, there are two types of motion in a an autonomous driving scene, namely motion of surrounding obstacles and the motion of the ego-vehicle. It is challenging to successfully classify the surrounding objects as moving or static when the camera reference itself is moving. In this case, even the objects that are not moving will be perceived as dynamic ones. Moving object detection implies two tasks that are performed jointly, namely, generic object detection which extracts specific classes such as pedestrians and vehicles. This is followed by motion classification, in which a classifier identifies the motion state of the object at hand among two classes, dynamic and static. Object detection and semantic segmentation has become a mature algorithm for automated driving \cite{siam2017deep} but motion segmentation is relatively an unexplored problem. Recent automated driving datasets \cite{yogamani2019woodscape} include moving object detection task. 

Recent CNN-based algorithms \cite{siam2018modnet,siam2018rtseg, ravi2018real} explored the problem of end-to-end motion segmentation through usage of optical flow images providing the motion of the surrounding scene as a prior information for the network which learns to generate a binary mask of two classes, "Moving" and "Static". Motion segmentation can be integrated into a multi-task system along with segmentation and other tasks \cite{sistu2019neurall,chennupati2019multinet++}.
Motion Estimation also helps in better estimation of depth \cite{kumar2018monocular}. Motion information can be perceived implicitly through a stack of temporally sequential images \cite{yahiaoui2019fisheyemodnet}, or explicitly through an external motion map such as optical flow map\cite{rashed2019motion}. Implicit motion modelling is prone to failure due to increased complexity of the task as the network learns to model motion in addition to segmenting the interesting object classes. On the other hand, external optical flow encodes motion between two consecutive frames only without considering previous states of the scene which negatively affects the output in two ways. First, the network becomes sensitive to optical flow errors because motion is being understood from two frames only. Second, the output masks become temporally inconsistent as they are independent of each other across time and therefore masks are more prone to errors and rapid changing. Moreover, optical flow encodes two pieces of information, the motion of the surrounding obstacles, in addition to the motion of the ego-vehicle which results in significant motion vectors associated with the static objects as well. This leads to the incorrect perception of static objects as moving objects. Nevertheless, optical flow augmentation has proven to improve accuracy of MOD compared to motion modelling from single color images due to understanding the motion across two sequential images such as in \cite{Valada_2017_IROS,siam2018modnet,siam2018real}. These results raised our question of how a CNN architecture would behave if it considers the previous states of the surrounding obstacles. 

In this work, we explore the benefit of leveraging temporal information through implementation of time-aware CNN architecture in addition to explicit motion modelling through optical flow maps. Moreover, we focus on real-time performance due to the nature of the autonomous driving task \cite{kiran2011parallelizing, briot2018analysis, siam2018rtseg}.
To summarize, the contributions of this work include:
\begin{itemize}
    \item Implementation of a novel CNN architecture for MOD utilizing spatio-temporal information. Our model combines both explicit and implicit motion modelling for maximum performance, and unlike previous baselines it ensures temporal consistency between successive frames.
    \item Construction of real-time performance network which significantly outperforms state-of-the art approaches and becomes suitable for time-critical applications such as the autonomous driving.
    \item Ablation study for various time-aware architectures for maximum performance in terms of accuracy, speed and temporal consistency.
    % \item Empirical evaluation for usage of explicit motion modelling compared to implicit motion perception through time-aware networks.
\end{itemize}

The rest of the paper is organized as follows: a review of the related work is presented in Section \ref{sec:relatedWork}. Our methodology including the dataset preparation and the used network architectures is detailed in Section \ref{sec:method}. Experimental setup and final results are illustrated in Section \ref{sec:experiments}. Finally, Section \ref{sec:conclusions} concludes the paper.

\section{Related Work}\label{sec:relatedWork}

% \begin{figure}
%   \centering
% %   \fbox{\rule[-.5cm]{0cm}{4cm} \rule[-.5cm]{4cm}{0cm}}
% \includegraphics[width=1\textwidth]{include/images/unfolded.png}
% %     \vspace{-1cm}
%   \caption{\textbf{Top:} Our baseline architecture based on \cite{zhang2018shufflenet}. \textbf{Bottom: } Our proposed architecture after optical flow and time augmentation.}
%   \label{fig:arch}
% \end{figure}

Classical approaches based on geometrical understanding of the scene such as \cite{menze2015object} have been suggested for motion masks estimation. Wehrwein et al. \cite{wehrwein2017video} introduced some assumptions to model the motion of the background as homography. This approach is very difficult to be used in AD due to the limited assumptions which causes errors such as camera translations assumptions. Classical methods generally provide less performance than deep learning methods in addition to the need to use complicated pipelines which introduce higher complexity in the algorithm. For instance, Menze et al. \cite{menze2015object} runs at 50 minutes per frame which is not suitable for AD.

Generic foreground segmentation using optical flow has been proposed by Jain et. al.\cite{jain2017fusionseg}, however it does not provide information about the state of each obstacle whether it is moving or static.
In \cite{drayer2016object,tokmakov2017learning} video object segmentation has been studied, however these networks are not practical for AD due to high complexity where they depend on R-CNN as in \cite{drayer2016object}, and DeepLab as in \cite{tokmakov2017learning} which run in 8 fps. Motion segmentation using deep network architectures has been explored by Siam et al. \cite{siam2018modnet,siam2018real}. These networks rely only on explicit motion information from optical flow which makes them sensitive to the optical flow estimation errors. Fisheye MOD has been studied in \cite{yahiaoui2019fisheyemodnet} using publicly available fisheye dataset \cite{yogamani2019woodscape} proving the importance temporally sequential images in MOD.

LiDAR sensors have been explored for MOD as as well, where most of LiDAR-based methods used clustering approaches such as \cite {dewan2016motion} to predict the motion of points using methods such as RANSAC, and then clustering takes place on the object level. Deep learning has been explored as well for such problem. 3D convolution is used in \cite {li20173d} to detect vehicles. Other methods projected the 3D points on images to make use of 2D convolutions on the image 2D space instead of 3D space \cite {li2016vehicle}. Low-illumination MOD has been explored by \cite{Rashed_2019_ICCV_Workshops} where optical flow has been utilized from both camera and LiDAR sensors demonstrating the importance of explicit motion modelling. Recent work \cite{dewan2017deep} predicts motion of objects from two input Lidar scans. This method uses implicit learning for motion information through two sequential Lidar scans and does not discuss the impact of time-aware networks which motivates our work to towards this study.

In this work, we aim to provide a real-time cost-effective approach for the autonomous driving application. Unlike LiDAR sensors, camera sensors have high efficiency compared to their cost. Thus, we focus on camera-based motion segmentation approaches.
\section{Methodology}\label{sec:method}

% \begin{figure}
%   \centering
% %   \fbox{\rule[-.5cm]{0cm}{4cm} \rule[-.5cm]{4cm}{0cm}}
% \includegraphics[width=1\textwidth]{include/images/architecture_2.png}
% %     \vspace{-1cm}
%   \caption{\textbf{Top:} Our baseline architecture based on \cite{zhang2018shufflenet}. \textbf{Bottom: } Our proposed architecture after optical flow and time augmentation.}
%   \label{fig:arch}
% \end{figure}

\begin{figure}
  \centering
%   \fbox{\rule[-.5cm]{0cm}{4cm} \rule[-.5cm]{4cm}{0cm}}
\includegraphics[width=1\textwidth]{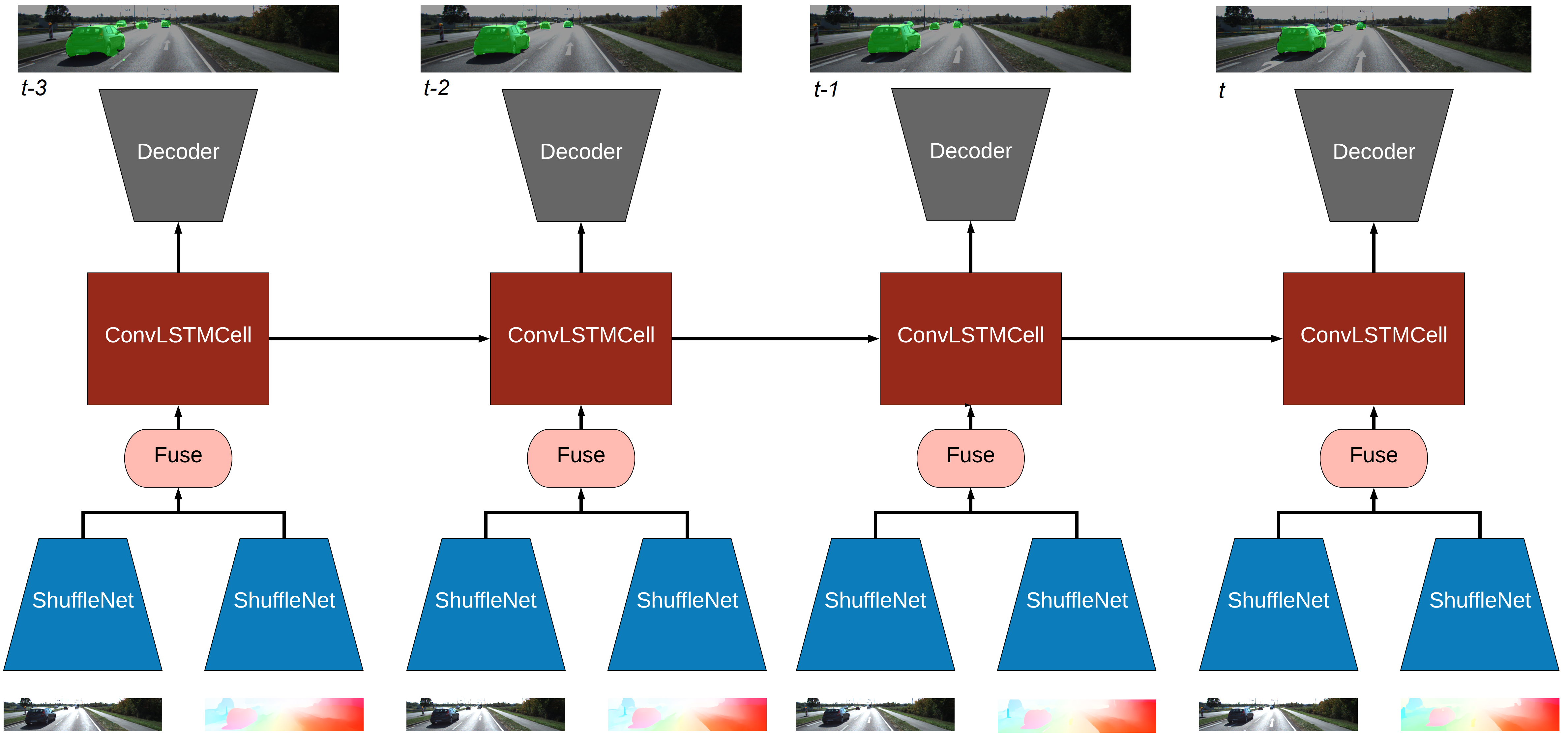}
%     \vspace{-1cm}
  \caption{Detailed architecture for our approach demonstrating how temporal information is used within the unfolded ConvLSTM cell.}
  \label{fig:unfolded}
\end{figure}

In this section we discuss dataset preparation, and detail the proposed architectures for our experiments.

\subsection{Dataset Preparation} \label{subsec:baseline}
% \noindent\textbf{Annotations Generation:}
\subsubsection{Annotations Generation}
To be able to train our deep model for good generalization, a large dataset including motion masks annotations is needed. There is huge limitation in MOD public datasets. For instance, Siam et al. \cite{siam2018modnet} provides 1300 images only with weak annotation for MOD task. On the other hand, 255 annotated frames are provided by Valada et al. \cite{Valada_2017_IROS} on KITTI dataset, while 3475 frames are provided on Cityscapes \cite{Cordts2016Cityscapes}. Cityscapes does not provide information about 3D motion which makes it not suitable for us to extend the dataset. Behley et al. \cite{behley2019iccv} provides MOD annotations for 3D point clouds only, but not for dense pixels. 
Due to this limitation in datasets, we build our own MOD dataset. We make use of the method in \cite{siam2018modnet} to generate motion masks from KITTI in order to extend the KittiMoSeg dataset. The bounding boxes provided by KITTI dataset are projected on 2D images while the tracking information is used 3D velodyne coordinate system to compute the velocity of each object compared to the ego-vehicle. The ego-vehicle velocity is provided via GPS annotation which allows us to compute relative speed between the ego-vehicle and the surrounding objects. We compare the relative speed to predefined thresholds to classify whether the obstacle is moving or static. This method is used to generate semi-automatic annotation for 12k images and then manual filtering is performed for fine tuning.
 
% \noindent\textbf{Color Signal:}
\subsubsection{Color Signal}
KITTI dataset\cite{Geiger2012CVPR} is used as it provides temporal sequences which we exploit to develop our time-aware architecture. Moreover, it has 3D bounding box and GPS coordinates annotations which we use to build our motion mask annotations. The upper part of the image is cropped as it is mainly dominated by sky and has no information about moving objects. The final resolution we train our network with is 256x1224. 

% \begin{figure}[t!]
% \centering
%     \includegraphics[width=.48\textwidth]{images/summaries/night_generation.png}
%     \caption{\textcolor{black}{An example of our different night generation methods, \textbf{Top to Bottom:} Input KITTI Image, Neural Style Transfer \cite{li2018closed}, CycleGAN\cite{Zhu-ICCV-2017}, UNIT\cite{liu2017unsupervised}}}
%     \vspace{-0.4cm}
% \label{nightImages}
% \end{figure}%

% \noindent\textbf{Motion Signal:} \label{subsec:motion}
\subsubsection{Motion Signal}
Motion can be either implicitly learned from temporally sequential frames, or provided explicitly to the system through an input motion map, as for example optical flow maps. In our approach, we develop a hybrid model that combines both methods together for maximum performance. FlowNet\cite{Ilg2016FlowNet2E} architecture is leveraged to generate the flow maps. We use color-wheel representation as it was observed to provide the best performance for encoding both magnitude and direction of motion. This is consistent with the observation of \cite{siam2018modnet,Rashed2019OpticalFA}.

\subsection{Network Architecture}

\begin{table}
  \caption{Quantitative comparison between different network architectures for MOD.}
  \label{tab:accuracy}
  \centering
  \begin{tabular}{lll}
    \toprule
    % \multicolumn{2}{c}{Part}                   \\
    \cmidrule(r){1-2}
    Experiment     & mIoU     & Moving IoU  \\
    \midrule
    RGB & 65.6  & 32.7      \\
    RGB+Flow & 74.24 & 49.36     \\
    RGB+Flow frame stacking & 63  & 27     \\
    RGB+Flow 3D Convolution & 64.3  & 29.8     \\
    RGB+Flow - LSTM (Early) & 73.5 & 48    \\
    RGB+Flow - LSTM (Late) & 69.2  & 39.3     \\
    RGB+Flow - LSTM (Multistage-2-filters)  & 73.7  & 48.5     \\
    \cmidrule(r){1-3}
    RGB+Flow - GRU (Multistage) & 75  & 50.9    \\
    RGB+Flow - LSTM (Multistage) & \textbf{76.3}  & \textbf{53.3}  \\
    
    \bottomrule
  \end{tabular}
\end{table}

In this section, we detail our baseline architecture, discuss how to use motion information and how to maximize the benefit of temporal sequences for MOD.

%\vspace{1mm}
\subsubsection{Baseline Architecture} 
Our baseline model is based on \cite{gamal2018shuffleseg}. The network is composed of an encoder-decoder architecture where feature extraction is performed by \cite{zhang2018shufflenet} reducing computational cost at high level of accuracy which is perfect for AD application. The decoder is based on \cite{long2015fully} which consists of 3 transposed convolution layers that are used to upsample the low resolution feature maps to the original image resolution. The network is trained to predict two classes, i.e, Moving and Non-Moving. There is huge imbalance between the two classes because of the background pixels which are considered static pixels as well. Weighted cross entropy is used to tackle the problem. The baseline architecture is used to evaluate the impact of RGB images only on MOD as illustrated in Figure \ref{fig:arch}

\subsubsection{Motion Augmentation}
As demonstrated by \cite{siam2018modnet}, explicit motion modelling through optical flow provides more accuracy than implicit learning of motion cues through temporal sequences. This is done through a 2-stream mid-fusion architecture which combines the processed feature maps from both network branches. It has been shown by \cite{rashed2019motion,Rashed2019OpticalFA} that mid-fusion architecture outperforms early-fusion which is based on raw data fusion before feature extraction. Feature-level fusion provides maximum accuracy at the cost of network complexity as the number of weights in encoder part is doubled. We adopt this approach for comparative study where semantic information is combined with motion information as illustrated in Figure \ref{fig:arch} and we demonstrate the impact on real-time performance.

\subsubsection{Time Augmentation} 
The main contribution of this work is to study the impact of including temporal information for MOD. For that purpose, we build upon the mid-fusion network and provide empirical study for various time-aware network architectures. We discuss the effect of using Frame stacking, 3D convolution, ConvLSTM\cite{xingjian2015convolutional}, and simpler GRU\cite{cho2014learning} layers which are time-dependent. For such experiments, we use a batch of 4 images as input as illustrated in Figure \ref{fig:unfolded} which explains how ConvLSTM is unfolded utilizing sequence of images for MOD. We design three network architectures leveraging ConvLSTM layers and provide empirical study to demonstrate their benefit for MOD.

\textbf{Early-LSTM:} In this case, we refer to the usage of ConvLSTM layer in each encoder separately, then fusion is done on the processed information.

\textbf{Late-LSTM:} In this case, we refer to the usage of ConvLSTM at the decision level before softmax layer where the network learns to use time information before the final classification is done. 

\textbf{Multistage-LSTM:} We implement several ConvLSTM layers across the network at 3 different stages as illustrated in Figure \ref{fig:arch} (Bottom). Finally, by "Multistage-2-filters" we refer to using 1x1 convolutional layers which squeezes the depth of the feature maps to \textit{num\_classes} and then apply ConvLSTM to the output channels.

\section{Experiments} \label{sec:experiments}

\begin{table}
  \caption{Quantitative results on KITTI-Motion\cite{Valada_2017_IROS} dataset in terms of mean intersection over union (mIoU) and running frames per second (fps) compared to state-of-the-art methods.}
  \label{tab:comparison}
  \centering
  \begin{tabular}{lll}
    \toprule
    % \multicolumn{2}{c}{Part}                   \\
    \cmidrule(r){1-2}
    Experiment     & mIoU     & fps \\
    \midrule
    CRF-M\cite{dinesh2015semantic} & 77.9  & 0.004     \\
    MODNet\cite{siam2018modnet} & 72  & 6     \\
    SmSNet\cite{Valada_2017_IROS} & \textbf{84.1} & 7      \\
    RTMotSeg\cite{siam2018real} & 68.8  & \textbf{25}     \\
    \cmidrule(r){1-3}
    RST-MODNet-GRU (ours) & 82.5 & 23    \\
    RST-MODNet-LSTM (ours) & 83.7 & 21   \\
    \bottomrule
  \end{tabular}
\end{table}

% \begin{figure}
%   \centering
%   \fbox{\rule[-.5cm]{0cm}{4cm} \rule[-.5cm]{4cm}{0cm}}
%   \caption{Sample figure caption.}
% \end{figure}

% \begin{table}
%   \caption{Quantitative comparison between different network architectures for MOD.}
%   \label{tab:accuracy}
%   \centering
%   \begin{tabular}{lll}
%     \toprule
%     \multicolumn{2}{c}{Part}                   \\
%     \cmidrule(r){1-2}
%     Experiment     & mIoU     & Moving IoU \\
%     \midrule
%     RGB & 65.6  & 32.7     \\
%     RGB+Flow & 74.24 & 49.36      \\
%     RGB+Flow frame stacking & 63  & 27     \\
%     RGB+Flow 3D Convolution & 64.3  & 29.8     \\
%     RGB+Flow - LSTM (Early) & 73.5 & 48    \\
%     RGB+Flow - LSTM (Multistage) & \textbf{76.3}  & \textbf{53.3} \\
%     RGB+Flow - LSTM (Multistage-2-filters)  & 73.7  & 48.5     \\
%     RGB+Flow - LSTM (Late) & 69.2  & 39.3     \\
%     RGB+Flow - GRU (Mid) & 75  & 50.9     \\
%     \bottomrule
%   \end{tabular}
% \end{table}

\subsection{Experimental Setup} \label{setup}
In our experiments, ShuffleSeg \cite{gamal2018shuffleseg} model was used with pre-trained ShuffleNet encoder on Cityscapes dataset for semantic segmentation except for the 3D Convolution experiment as we randomly initialized the encoder weights. For the decoder part,  FCN8s decoder has been utilized with randomly initialized weights. L2 regularization with weight decay rate of $5e^{-4}$ and Batch Normalization are incorporated. We trained all our models end-to-end with weighted binary cross-entropy loss for 200 epochs using 4 sequential frames. Adam optimizer is used with learning rate of $1e^{-4}$.
For frame stacking experiments we modified the depth of the first convolutional layer filters to match the input by replicating the filters in the depth dimension to utilize Cityscapes weights instead of randomly initializing them.

%%%% One Image to reduce size
\begin{figure*}[t!]
% \captionsetup[subfigure]{labelformat=empty}
\centering
% \begin{subfigure}{\textwidth}
    \includegraphics[width=\textwidth]{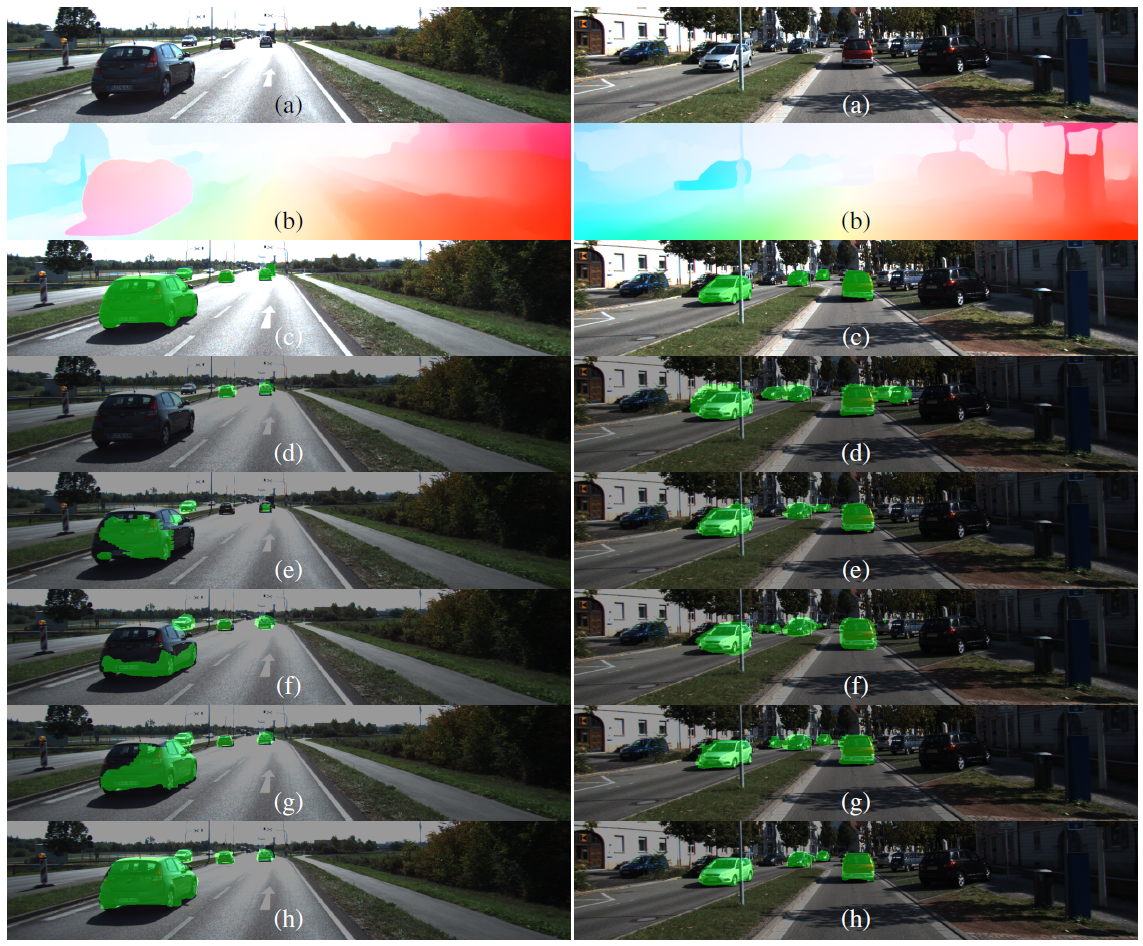}
    % \vspace{-1cm}
\caption{Qualitative comparison of our algorithm on two samples of KITTI dataset each sample is represented in a column. \textbf{(a),(b)} show the input RGB images and flow images. \textbf{(c)} shows ground truth. \textbf{(d)} shows RGB-only output. \textbf{(e)} shows RGB+Flow output. \textbf{(f)} output of RGB+Flow-LSTM(Late). \textbf{(g)} shows output of RGB+Flow-GRU(Multistage). \textbf{(h)} shows output of the proposed architecture RGB+Flow-LSTM(Multistage).}
%     \vspace{-0.4cm}
    \label{fig:qualitative_1}
% \end{subfigure}%
\end{figure*}

\subsection{Experimental Results} \label{sec:results}
 We provide a table of quantitative results for our approach evaluated on KITTI dataset and a table for comparison with state-of-the-art baselines on KITTI-Motion dataset \cite{Valada_2017_IROS}. Qualitative evaluation is illustrated in Figure \ref{fig:qualitative_1}. Table \ref{tab:accuracy} demonstrates our results in terms of mIoU on both classes in addition to IoU for the Moving class. RGB-only experiment result is used as a comparison reference where color information only is used for classifying MOD without using either implicit or explicit motion modelling. Significant improvement for 17\% in moving class IoU has been observed after fusion with optical flow, which is consistent with previous conclusions in \cite{siam2018modnet,siam2018real}. Naive frame stacking showed inability of the network to benefit from the temporal information embedded into the sequence of images while 3D convolution increased the network complexity dramatically which made it not suitable for embedded platform for autonomous driving application. For that reason we focus our experiments on usage of ConvLSTM layers where we provide an empirical evaluation to figure out the best architecture utilizing ConvLSTM. Three architectures are implemented using ConvLSTM. Early and Late LSTM show improved performance over the baseline RGB, however they perform very close to standard two-stream RGB+Flow which means the information is not fully utilized. This encourages the implementation of a mid-level ConvLSTM-based network that captures motion in multiple stages. Using this architecture, we obtain absolute improvement of 4\% in accuracy and relative improvement of 8\% over RGB+Flow showing the benefit of our approach. We provide two versions of our multistage architecture comparing ConvLSTM and GRU. We observe very close performance in terms of accuracy with slightly higher running rate using GRU which is expected due to simpler internal architecture.
 
 Table \ref{tab:comparison} shows a comparison between our approach and state-of-the-art baseline methods. For fair comparison of model speed, we run all the tests on our Titan X Pascal GPU using the same input resolution in \cite{Valada_2017_IROS}. RTMotSeg\cite{siam2018real} has two models, one of which is using LiDAR point cloud in a post processing step to minimize false positives. We report the model which does not use LiDAR sensor as we mainly focus on camera-based solutions. It is shown that our method is on par with the baseline methods where we provide almost the same accuracy as SMSNet\cite{Valada_2017_IROS}, however at almost double the inference speed using both our multistage time-aware architectures which makes them more suitable for embedded platform for autonomous driving applications.
 
 Figure \ref{fig:qualitative_1} shows qualitative results on two KITTI samples demonstrating the benefit of using time-aware networks for MOD where each column represents a sample. (a),(b) show the input RGB and optical flow inputs. (c) shows the motion mask ground truth. (d) shows inability of CNN to understand motion information from color images only without sequence of images or optical flow maps. (e) shows improvement over RGB-only due to fusion with optical flow which encodes motion of the scene. (f) shows the output of RGB+Flow after adding LSTM layer before softmax layer (Late) which demonstrates the improvement over RGB-only as illustrated in Table \ref{tab:accuracy}. However, the network is still unable to completely utilize the motion information embedded within the sequence of images. (g),(h) show the output of our multistage models,namely RGB+Flow-GRU in (g) and RGB+Flow-LSTM in (h). Results visually confirm the benefit of our algorithm through implementation of multistage time-aware layers where motion information is fully exploited. 
 
\begin{figure*}[t!]
\captionsetup[subfigure]{labelformat=empty}
\centering
% \begin{subfigure}{\textwidth}
    \includegraphics[width=\textwidth]{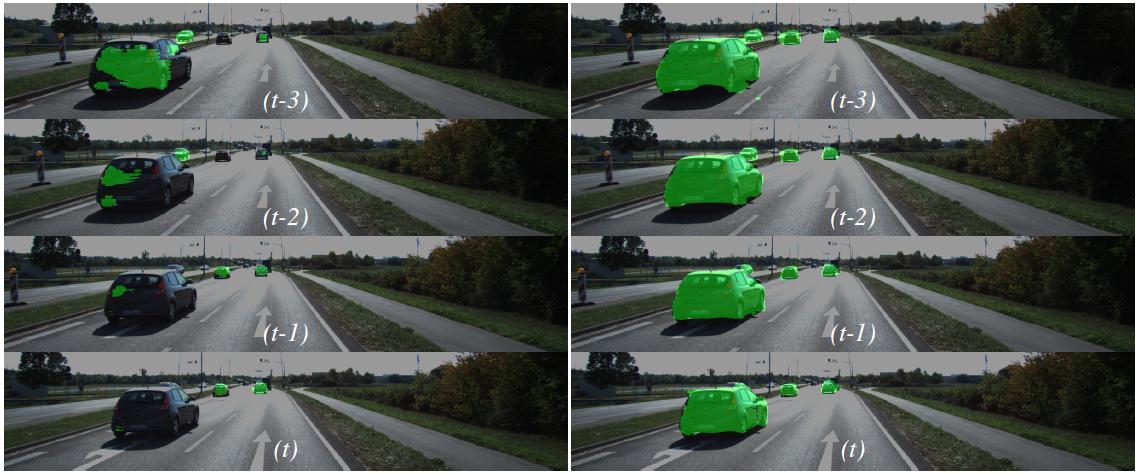}
    % \vspace{-1cm}
    % \caption{\textcolor{white}{\textit{(t-3)}}}
 %   \caption{\textcolor{white}{KITTI input image}\newline}
 \caption{Qualitative evaluation demonstrating the temporal consistency obtained from our approach on the right column compared to RGB+Flow on the left column as previous baselines. \\}
 \label{fig:sequence_1}
% \end{subfigure}%
\end{figure*}

\begin{figure*}[t!]
\captionsetup[subfigure]{labelformat=empty}
\centering
% \begin{subfigure}{\textwidth}
    \includegraphics[width=\textwidth]{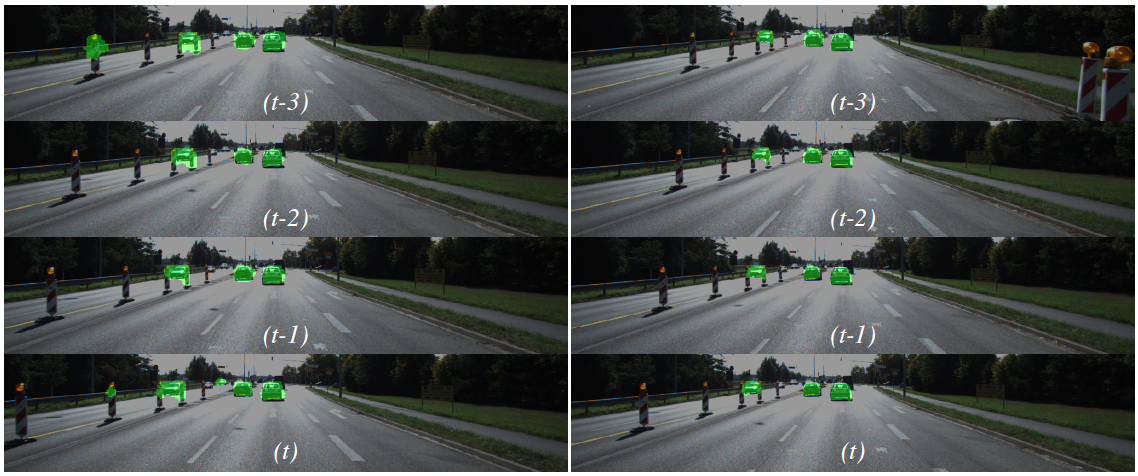}
    % \vspace{-1cm}
    % \caption{\textcolor{white}{\textit{(t-3)}}}
    \caption{Qualitative evaluation demonstrating the temporal consistency obtained from our approach on the right column compared to RGB+Flow on the left column as previous baselines.}
    \label{fig:sequence_2}
% \end{subfigure}%
\end{figure*}

 Figures \ref{fig:sequence_1},\ref{fig:sequence_2} show the advantage of our approach across time where relationship between sequential images has been modelled within the network. Each figure represents a sample sequence from KITTI data where the left column represents the output of RGB+Flow while the right column shows the impact of our algorithm considering time information. On the left column, the furthest car in time \textit{t-3} has been segmented correctly then accuracy is lost in \textit{t-2} and obtained again in \textit{t-1}. This is also shown on the close car on the left where the mask is sensitive to optical flow map. On the other hand, the right column shows temporally consistent motion masks after the addition of our multistage-LSTM layers within the network. The same conclusion is obtained from Figure \ref{fig:sequence_2}, where these results demonstrate the improved performance in Table \ref{tab:accuracy}.

\section{Conclusions} \label{sec:conclusions}
In this paper, we propose a novel method for moving object detection which balances between high accuracy and high computational efficiency. Our proposed method exploits both external motion modelling and time-aware architectures to maximize benefit from temporal motion information. An ablation study is provided for various time-aware architectures to evaluate the impact of our approach on MOD. The algorithm is evaluated on KITTI and KITTI-Motion datasets against state-of-the-art baselines. We obtain 8\% relative improvement in accuracy after augmentation of time-aware layers. Competitive results are demonstrated in terms of accuracy compared to state-of-the-art SMSNet model at three times the inference speed which makes our algorithm more suitable for autonomous driving application. 

\medskip
\clearpage % Remove me if the layout gets fucked up!
% % %%%%%%%%%%%%%%%%%%%%%%%%%%%%%%%%%%%%
% % References
% % %%%%%%%%%%%%%%%%%%%%%%%%%%%%%%%%%%%%
{\small
\bibliographystyle{unsrt}
\bibliography{references}
}

\end{document}